\documentclass[letterpaper]{article} 
\usepackage{aaai23}  
\usepackage{times}  
\usepackage{helvet}  
\usepackage{courier}  
\usepackage[hyphens]{url}  
\usepackage{graphicx} 
\usepackage{natbib}  
\usepackage{caption} 
\usepackage{algorithm}
\usepackage{algorithmic}
\usepackage{amsmath,booktabs,subcaption,siunitx,multirow,amsfonts,marvosym}
\usepackage{newfloat}
\usepackage{listings}
\DeclareCaptionStyle{ruled}{labelfont=normalfont,labelsep=colon,strut=off} 
\floatstyle{ruled}
\newfloat{listing}{tb}{lst}{}
\floatname{listing}{Listing}
\title{Learning to Select Prototypical Parts for Interpretable Sequential Data Modeling}
\author {
    Yifei Zhang,
    Neng Gao\footnote{Corresponding author.}, 
    Cunqing Ma 
}
\affiliations {
    Institute of Information Engineering, Chinese Academy of Sciences, Beijing, China \\
    \{zhangyifei, gaoneng, macunqing\}@iie.ac.cn
}

\begin{document}

\maketitle

\begin{abstract}
Prototype-based interpretability methods provide intuitive explanations of model prediction by comparing samples to a reference set of memorized exemplars or typical representatives in terms of similarity.
In the field of sequential data modeling, similarity calculations of prototypes are usually based on encoded representation vectors.
However, due to highly recursive functions, there is usually a non-negligible disparity between the prototype-based explanations and the original input.
In this work, we propose a Self-Explaining Selective Model (SESM) that uses a linear combination of prototypical concepts to explain its own predictions.
The model employs the idea of case-based reasoning by selecting sub-sequences of the input that mostly activate different concepts as prototypical parts, which users can compare to sub-sequences selected from different example inputs to understand model decisions.
For better interpretability, we design multiple constraints including diversity, stability, and locality as training objectives.
Extensive experiments in different domains demonstrate that our method exhibits promising interpretability and competitive accuracy.
\end{abstract}

\section{Introduction}

Deep neural networks have been widely employed for analyzing sequential data in real world, such as electrocardiogram (ECG), event streams, and natural language text.
State-of-the-art methods generally leverage CNN, RNN, or other hybrid networks for sequence modeling.
To meet the growing demand on explaining model predictions from black-box deep neural networks, more and more researchers resort to interpretable machine learning (IML) methods.

IML methods for sequential data can be roughly divided into three categories, including \textit{post-hoc methods}, \textit{attention-based methods}, and \textit{prototype-based methods} \cite{DBLP:journals/pieee/SamekMLAM21}.
Post-hoc methods \cite{DBLP:conf/emnlp/JacoviSG18} are the only options to provide \textit{a posteriori} explanations for trained models by assessing the impact of different input features and approximating the decision-making process.
Attention-based methods and prototype-based methods explicitly reveal some significant information for model prediction in different ways, resulting in model-intrinsic interpretability.
Specifically, attention-based methods \cite{DBLP:conf/cikm/WangLLW19} introduce additional parameters and specific model structure to weigh the importance of sequence elements, namely attention mechanism, for giving insight into models.
Prototype-based methods \cite{DBLP:conf/aaai/LiLCR18}, derived from case-based reasoning \cite{DBLP:conf/nips/KimRS14}, design machine learning models that select or create a set of representative instances as prototypes (also named as concepts).
The model tries to find several prototypes that closely resemble an input for prediction.
By inspecting the selected prototypes, the explanations are more intuitively understandable for laypersons.

However, recent efforts are often faced with the following challenges.
Post-hoc explanations are shown to be incorrect or incomplete, since the approximation may not always reflects the real model structure \cite{DBLP:conf/ijcai/LaugelLMRD19,DBLP:conf/aies/LakkarajuB20}.
The interpretability of attention mechanism remains controversial \cite{DBLP:conf/naacl/JainW19,DBLP:conf/emnlp/WiegreffeP19}.
Although attention mechanism has been widely applied and helps models attend to more significant elements or features, some have questioned that attention-based explanations are unreliable or unfaithful \cite{DBLP:conf/acl/SerranoS19,DBLP:conf/kdd/BaiLZLBW21}.
Existing prototype-based methods mostly learn a fixed set of representative vectors as prototypes, and an input is represented by one or more prototypes as explanation in practice.
For modeling sequences with rich information, state-of-the-art methods \cite{DBLP:conf/kdd/MingXQR19,DBLP:conf/nips/ChenLTBRS19,DBLP:journals/jmlr/ArikP20} need to maintain a rather large number of prototypes to achieve reliable performance \cite{DBLP:journals/corr/abs-2007-01777}.
Moreover, the similarity-based measurements for finding prototypes are applied on hidden representations generated by encoders with highly recursive functions, thus the direct relation between prototypes and the input sequence might still be hard to understand for human.
Overall analysis of the provided prototypes as well as the input sequence is required, which deviates from the interpretability assumption to some extent.

\begin{figure}[tb]
	\centering
	\begin{subfigure}[b]{\linewidth}
		\centering
		\includegraphics[width=0.8\linewidth]{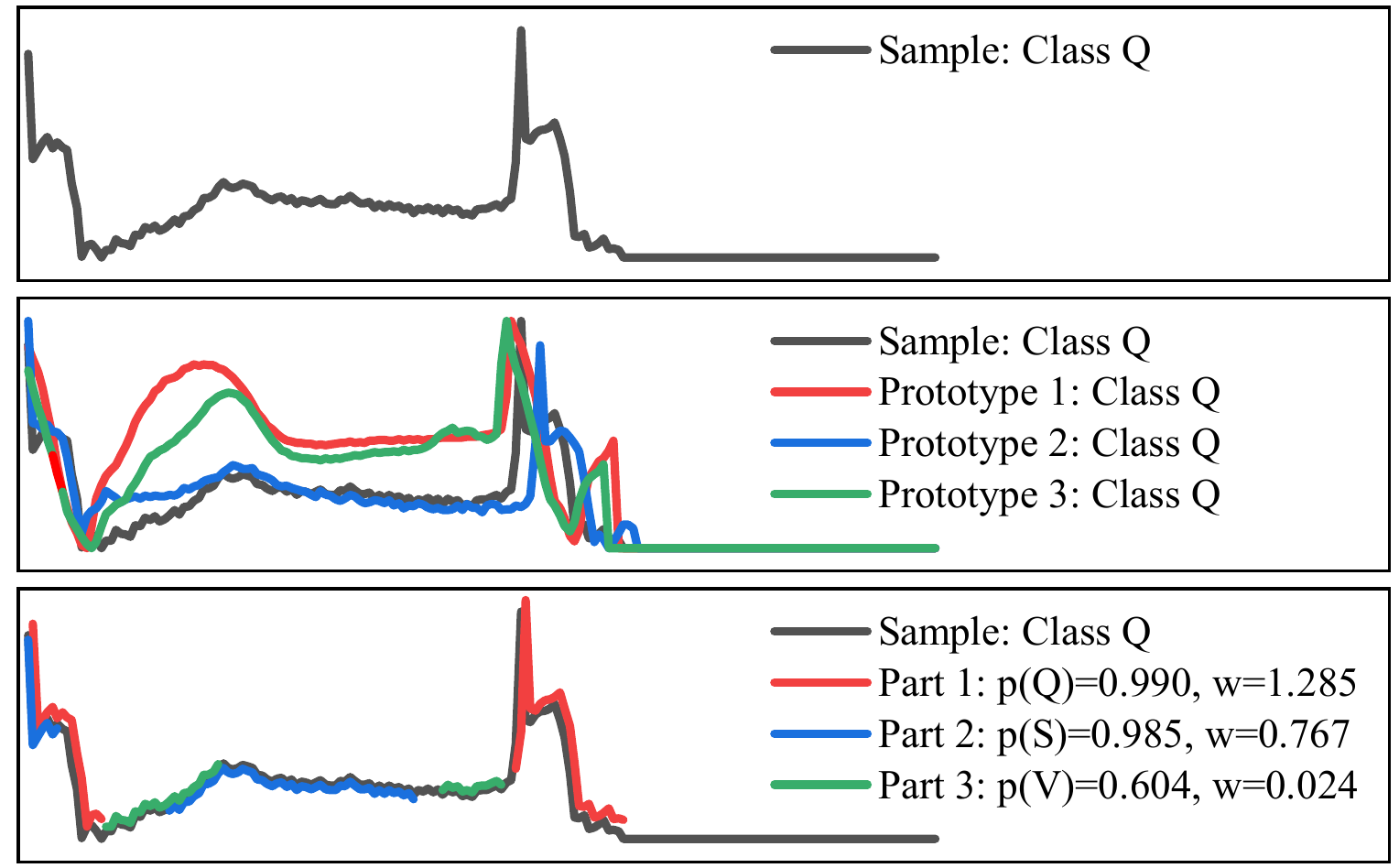}
		\caption{ECG signal classification. Top: signal sample. Center: prototypes of ProSeNet. Bottom: prototypical parts of proposed SESM.}
	\end{subfigure}
	\hfill
	\begin{subfigure}[b]{\linewidth}
		\centering
		\includegraphics[width=\linewidth]{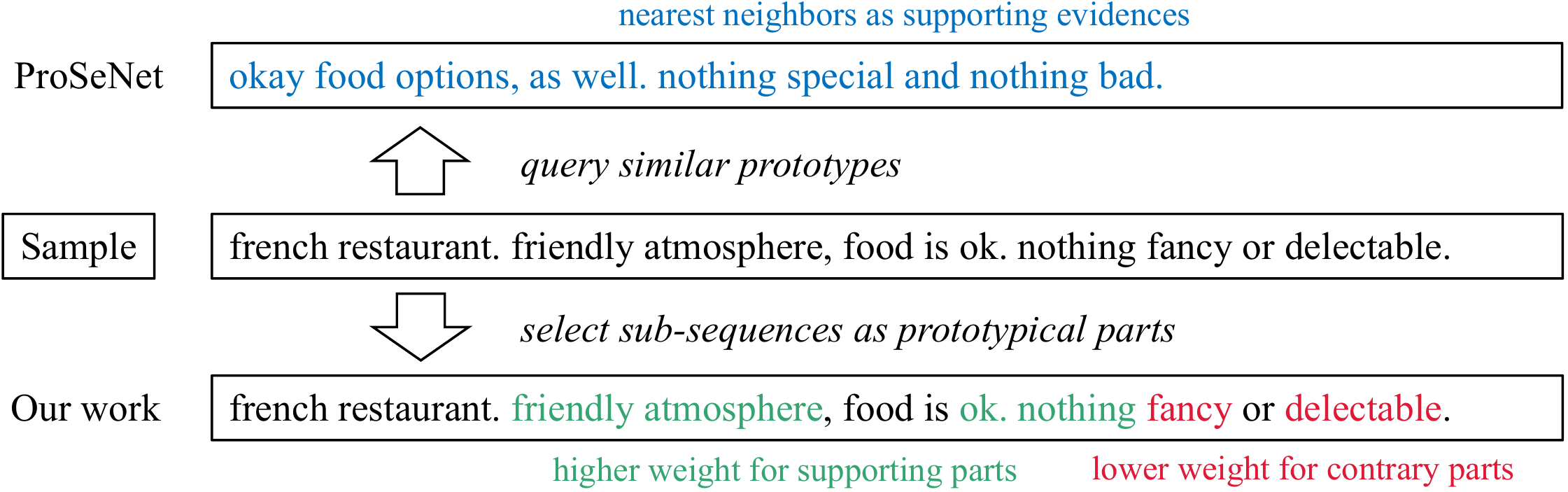}
		\caption{Review semantic classification. Color distinguishes words from different sub-sequences.}
	\end{subfigure}
	\caption{Difference between similarity-based prototypes and the proposed self-selective prototypical parts. }
	\label{fig:idea}
\end{figure}

In this work, we suggest enhancing the prototype-based interpretability for sequential data with self-selective prototypical parts, and design a \underline{S}elf-\underline{E}xplaining \underline{S}elective \underline{M}odel (SESM).
Rather than maintaining a relatively large amount of instances and find the nearest neighbor for explanation, SESM explains by selecting sub-sequences that represent disentangled concepts of the input sequence as prototypical parts.
Motivated by the general framework of the self-explaining neural network (SENN) \cite{DBLP:conf/nips/Alvarez-MelisJ18}, SESM leverages a modified multi-head self-attention mechanism as conceptizer to select prototypical parts, where each head tends to select the sub-sequence of the input that mostly activates a specific concept.
Then, the prototypical parts of an input are encoded separately and aggregated linearly as the final modeling result.
Note that since the concepts are represented in the form of sub-sequences, they can be encoded by any existing sequence modeling methods and explained via prototyping (i.e., create a small set of sub-sequences from training set with similar concepts).
As instantiated in Figure~\ref{fig:idea}, the prototypical parts manifest as supporting or contrary evidences for model prediction according to their assigned weights, where the supporting evidence shows the decisive sub-sequences of the input, while the contrary evidence represents the part of the input that could lead the model to output a different class.
For better interpretability, we design several learning criteria to impose unsupervised disentanglement \cite{DBLP:journals/corr/abs-2103-11251} in the end-to-end training process, including diversity, stability, and locality, where diversity reduces redundancy of concepts, stability provides conceptual unity, and locality prevents prototype parts from degenerating into prototypes.
Hence, the model is inherently interpretable that can provide straightforward and brief explanations for model predictions.
Moreover, the selection of prototypical parts does not require storage of raw data from the training set, which could be a stepping stone towards GDPR compliance.
Our contributions of proposing SESM can be summarized as follows:

\begin{itemize} 
	\item SESM is an inherently interpretable sequence model, which selects sub-sequences representing disentangled concepts of an input as prototypical parts for explaining its own predicting process.

	\item SESM is end-to-end trainable with our designed learning criteria for the training process, which impose constraints on extracting straightforward and brief explanations, without memorizing a rather large amount of raw data from the training set.

	\item SESM shows comparable effectiveness and outperforms baseline methods in terms of interpretability based on our extensive experiments on various domains.
\end{itemize}

\section{Related Work}

Compared with post-hoc and attention-based explanations, prototype-based explanations are more easily understandable for human through case-based reasoning \cite{DBLP:conf/nips/KimRS14,DBLP:conf/nips/Alvarez-MelisJ18}.
PrototypeDL \cite{DBLP:conf/aaai/LiLCR18} maintains multiple representative vectors as prototypes, and the similarities between an input and the prototypes are concatenated as a representation vector for classification.
The prototype vectors are required to be as close as possible to a real training instance in latent space, so that humans may inspect the most similar training instances to the input in the latent space for explanation.
SENN \cite{DBLP:conf/nips/Alvarez-MelisJ18} automatically extracts representative concepts from inputs as prototypes, and samples instances that maximally activate each concept for explanation.
ProSeNet \cite{DBLP:conf/kdd/MingXQR19} provides prototype-based explanations for sequential data with the similar idea as PrototypeDL.
Instead of prototyping the entire sequences, SelfExplain \cite{DBLP:conf/emnlp/RajagopalBHT21} and SCNpro \cite{DBLP:conf/cikm/NiCCZSLZC21} prototype local segments of sequences for more accurate and understandable explanation.

Our work differs from existing work in the following aspects.
First, the framework of PrototypeDL and ProSeNet requires encoding the entire sequences for similarity-based prototype query.
The sequence-level encoding of input inevitably involves irrelevant or opposite elements, while SESM learns to select sub-sequences with disentangled concepts, which can provide more fine-grained explanations.
Second, the way SENN extracts concepts from the input is not interpretable, since the conceptizers are black-box neural networks.
Instead, SESM generates concepts with selective prototypical parts, which is a part of the original sequence to be immediately understandable.
SelfExplain also successfully addresses this issue by leveraging constituency-based parse trees to generate text spans of words and phrases from the input sentence as concepts, while at a cost of limited application scenario of neural language classification.
Third, the sub-sequences for prototyping claimed by SCNpro is actually local segments of continuous elements, while SESM works on actual sub-sequences that can be composed of discontiguous elements, which can help capture distant elements representing similar concepts with less amount of prototypes.
Moreover, we would like to stress that the the selective prototypical parts are created solely based on the selective actions, in order to completely eliminate the effect of discarded elements, and can be encoded as concepts with any existing sequence modeling methods.

\section{Methodology}

\subsection{Model Architecture} 

\begin{figure}[tb]
	\centering
	\includegraphics[width=\linewidth]{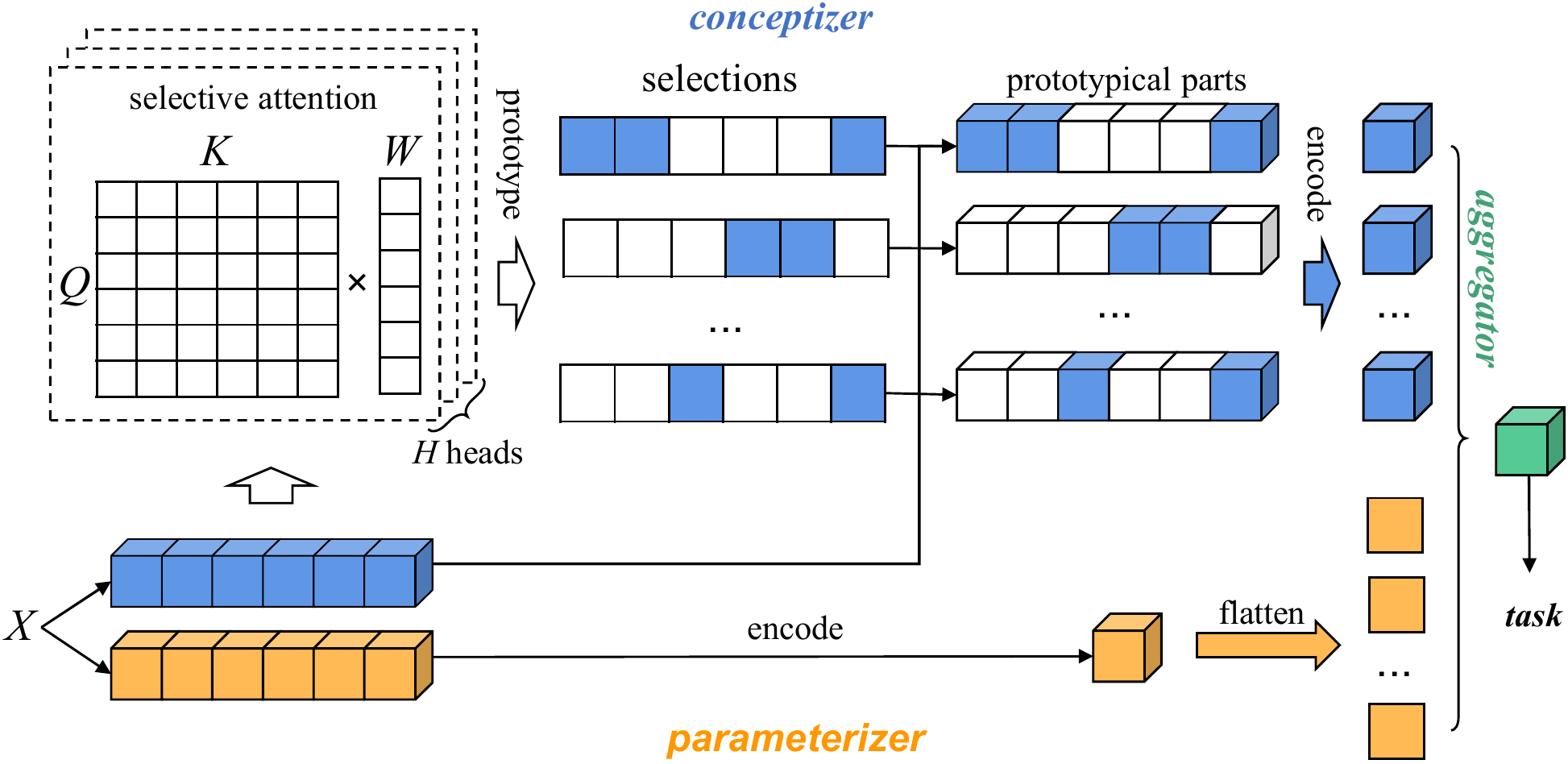}
	\caption{The overall architecture of our proposed SESM. Squares denote scalars and cubes denote vectors.}
	\label{fig:model}
\end{figure}

For inherently interpretability, SESM is comprised of three main modules following the self-explaining framework of SENN, including \textit{conceptizer}, \textit{parameterizer} and \textit{aggregator}.
The conceptizer selects multiple sub-sequences from the raw input as prototypical parts representing disentangled concepts, and the parameterizer determines which concepts are of greater importance for model prediction by assigning different weights.
Then, the aggregator linearly combines the prototypical parts to obtain the final representation of the input sequence.
The detailed architecture of SESM is shown in Figure~\ref{fig:model}.
Let $X=\{x_1,x_2,\cdots,x_N\}$ denote an input sequence with $N$ elements.
Next, we will formalize the detailed implementation of each module respectively.

The \textbf{conceptizer} $\mathcal{C}$ aims to create $H$ sub-sequences of $X$ with disentangled concepts $\{X_1,X_2,\cdots,X_H\}$.
Specifically, a vector of selective actions $s_h= \mathcal{C}_h(X) = \{0, 1\}^{N}$ is generated for each prototypical part $X_h$ based on the selective mechanism, indicating which elements $X_h$ is composed of.
By combining $H$ different selection vectors, the output of conceptizer can be denoted as a selection matrix $\mathcal{C}(X) = \{s_h\}^{H}_{h=1} = \{0, 1\}^{H\times N}$, where $s_{h,i}=1$ indicates that the $h$-th prototypical part includes the $i$-th element of $X$.

\citet{DBLP:conf/acl/GengWWQLT20} have proposed the selective self-attention network (SSAN), which is an ideal base method for realizing the function of our conceptizer with selective mechanism.
Given input sequence $X$, the SSAN embeds and projects $X$ into three $d_h$-dimensional matrices queries, keys and values $Q,K,V\in \mathbb{R}^{N\times d}$.
Then, SSAN applies the commonly used dot-product attention \cite{DBLP:conf/nips/VaswaniSPUJGKP17} to obtain the selective attention weights based on element pairs:
\begin{equation}\label{eq:ssan}
	\operatorname{ATT}(Q,K)=\operatorname{Gumbel-Sigmoid}\left(QK^{T}/\sqrt{d_h}\right),
\end{equation}
where the $\operatorname{Gumbel-Sigmoid}$ operation is a reparametrization trick for training non-differentiable model with selective operations, and the pair-wise attention is applied on $V$ as $\operatorname{ATT}(Q,K)\times V$.
By stacking multiple $Q,K,V$ combinations, namely multi-head attention, the final modeling result of $X$ is able to include different perspectives of information in parallel.

However, the operation of applying self-attention mechanism leads to relatively weak model interpretability.
The dot-product attention is based on entangled pair-wise information of elements in $Q$ and $K$, and is applied on $V$ through matrix multiplication, thus the attentive information cannot be directly associated to a small set of raw elements from the input sequence for intuitive explanation.

For better interpretability, we introduce an additional matrix for projection $W\in \mathbb{R}^{N\times 1}$ into Eq.~\ref{eq:ssan}.
By grouping the pair-wise attentions for each element in rows, $W$ squeezes the pair-wise attentions into element-wise attentions before binarized:
\begin{equation}\label{eq:sesm}
	\mathcal{C}_h(X)
	=\operatorname{Gumbel-Sigmoid}\left(QK^{T}W/\sqrt{d_h} \right).
\end{equation}
We can then extract sub-sequences as human-friendly prototypical parts for explanation with the element-wise selective attention.
After that, the concepts of an input is encoded according to the selected prototypical parts.
Let $Enc$ denote an arbitrary encoder, the encoded concept $c_h$ of the $h$-th prototypical part can be denoted as $c_h = Enc(X_h)=Enc(X \otimes s_h)$, where $\otimes$ denotes applying selection $s_h$ on input $X$.

The \textbf{parameterizer} $\mathcal{P}$ aims to decide the contributions of different concepts for model prediction.
According to SENN, the entries of parameters for interpretability should be non-negative.
The parameterizer models the entire sequence with stacked CNNs followed by an MLP for projection to weigh each prototypical part $X_h$, where the output of $\mathcal{P}(X)=\{p_h\}^H_{h=1}$ is an $H$-dimensional vector activated by Softplus.


The \textbf{aggregator} $\mathcal{G}$ aggregates the concepts of prototypical parts $c_h$ and their corresponding non-negative weights $p_h$ for an overall representation vector for downstream tasks.
The aggregation process is additive on the encoded concepts for interpretability.
As introduced above, a prototypical part $X_h$ is represented by a binary vector that indicates the selective action of each entry.
Thus, the aggregating result, i.e., the overall modeling result of SESM, can be formalized as:
\begin{equation}
	\operatorname{SESM}(X) = \mathcal{G}(\mathcal{P}(X)\mathcal{C}(X)) = \sum^H_{h=1} p_hc_h .
\end{equation}

\subsection{Learning Objective}

The learning objectives of SESM should meet the demand of both utility and interpretability.
Let $f_\text{task}(\cdot)$ denote a network for a downstream task (e.g., ECG signal classification, sentence semantic classification), and $y$ denote the corresponding ground-truth label in the training set.
The loss of utility is then $\mathcal{L}_\text{task}(f_\text{task}(\operatorname{SESM}(X)), y)$, e.g., negative log likelihood loss and cross-entropy loss.
For interpretability, we design three learning criteria as regularization for the conceptizer and parameterizer.
Next, we describe the proposed criteria along with the reasons to adopt them.

\subsubsection{Diversity}

The prototypical parts should comprise different elements of a sequence as well as representing different perspectives of information for disentanglement, in order to reduce redundancy (multiple prototypical parts represent similar concepts) and incompleteness (a single prototypical part does not cover all necessary elements) of explanation.
Accordingly, we design the diversity regularization to constrain that different prototypical parts select different elements in the input $X$.
Inspired by ProSeNet, we leverage L2 distance to design the loss function of diversity with threshold $d_{\min}=2$:
\begin{equation}
    \mathcal{L}_{d} = \sum^{H-1}_{i=1}\sum^{H}_{j=i+1} \biggl[\operatorname{RELU}\left( d_{\min} - \|s_i - s_j\|^2 \right)   \biggr] .
\end{equation}

\subsubsection{Stability}

For better disentanglement and stability, we require each head of the conceptizer to focus on a single concept.
Specifically, the encoded representations $c_h$ of prototypical parts selected by the same head of conceptizer $\mathcal{C}$ are similar.
The regularization is implemented by minimizing the pair-wise cosine distance of encoded concepts from the same head at batch level.
Formally, we have:
\begin{equation}
	\mathcal{L}_{s} = \sum^H_{h=1}\sum^{B-1}_{i=1}\sum^{B}_{j=i+1} \biggl[ 1 - cos(c^h_{i}, c^h_{j})\biggr],
\end{equation}
where $B$ denotes the batch size.

\subsubsection{Locality}

During experiment, we noticed that the conceptizer would occasionally fail, when some of the heads select all elements of the original sequence $X$ and the others select none.
To tackle this problem and further encourage diversity in the prototypical parts, we introduce locality loss to penalize heads that select an excessive amount of components:
\begin{equation}
	\mathcal{L}_{l} = \sum^H_{h=1} \frac{1}{N} \sum^N_{i=1}s_{h,i},
\end{equation}

To sum up, the overall learning objective is:
\begin{equation}
	\mathcal{L} = \mathcal{L}_{task} + \lambda_d\mathcal{L}_d + \lambda_s\mathcal{L}_s + \lambda_l\mathcal{L}_l,
\end{equation}
where $\lambda$ weighs each regularization term.

\subsection{Model Interpretability}

The interpretability of our proposed SESM is in two fold.
First, SESM selects several prototypical parts representing disentangled concepts and assign different weights to them, which illustrates what concepts a sequence is comprised of and the corresponding importance for model prediction, resulting in model-intrinsic interpretability via unsupervised disentanglement.
Then, the selected prototypical parts can be further interpreted via prototyping.
Prototype-based explanations can help users to understand model prediction through case-based reasoning.
By sampling the prototypical parts of instances in the training set that maximally activate a concept, users can intuitively inspect the concept by comparing the selected prototypical part from the input with the most influential prototypes, which provides fine-grained prototype-based interpretability.

\section{Experiments}

\subsection{Experimental Settings}

We compare SESM with multiple open source baselines, including black-box methods without interpretability LSTM and CNN, attention-based method named SAN \cite{DBLP:conf/iclr/LinFSYXZB17}, and methods with prototype-based interpretability ProSeNet and SelfExplain.
Since the original implementation of ProSeNet is not open-source, we migrate the reproduced TensorFlow version on github\footnote{https://github.com/rgmyr/tf-ProSeNet} for comparison.
SelfExplain relies on constituency-based parse trees generated by pre-trained neural
language models, thus it is only applied for comparison on natural language tasks.
For fairness, we do not compare SESM with the most recent work SCNpro, since it is not open source by far.
The codes for implementing SESM and our reproduction of ProSeNet will be released on Github\footnote{https://github.com/iiezyf/SESM}.
The experiments are conducted with \texttt{PyTorch} 1.8 on two NVIDIA Tesla V100 16GB GPUs.
The AdamW optimizer \cite{DBLP:conf/iclr/LoshchilovH19} is employed for all experiments, which improves the generalization performance of the commonly used Adam optimizer.
The experiment results are averaged on 10 runs with different random seeds.

We evaluate the effectiveness of our proposed SESM on tasks from various domains.
The accuracy and macro-averaged precision and recall are employed for measuring accuracy.
To the best of our knowledge, there is no standardized method for quantitatively assessing interpretability, thus we follow existing work \cite{DBLP:conf/aaai/LiLCR18,DBLP:conf/kdd/MingXQR19,DBLP:conf/cikm/NiCCZSLZC21} to highlight qualitative case studies, and migrate the area over perturbation curve (AOPC) \cite{DBLP:conf/naacl/Nguyen18,DBLP:conf/acl/ChenZJ20} for quantitative analysis.
AOPC is utilized as a counterfactual assessment of word-level explanations for text classification, by measuring the average change in the prediction probability on the predicted class after deleting top-scored words.
For prototype-based methods, we migrate AOPC to quantify the prototype-based interpretability of models by measuring the probability drop of the predicted class after deleting the most relevant prototypes (or prototypical parts in SESM):
\begin{equation}
	\operatorname{AOPC} = \frac{1}{H-1}\left< \sum^{H-1}_{h=1} f(\boldsymbol{x})-f(\boldsymbol{x}_{\backslash 1..h}) \right>,
\end{equation}
where $f(\boldsymbol{x}_{\backslash 1..h})$ is the probability on the predicted class \textit{without} the top $h$ relevant prototypes, and $\left<\cdot\right>$ denotes the average over samples.
A larger AOPC indicates that the model shifts its prediction more drastically and implies that the most contributing prototypes for the final prediction are of greater significance and less redundancy.

Next, we will briefly introduce the leveraged tasks and case studies about sequence modeling respectively.

\subsubsection{Task~1: ECG Signal Classification}

We first evaluate SESM on real-valued ECG signal classification task using the pre-processed MIT-BIH Arrhythmia dataset\footnote{https://www.kaggle.com/shayanfazeli/heartbeat} as existing work \cite{DBLP:conf/kdd/MingXQR19,DBLP:conf/cikm/NiCCZSLZC21}.
The MIT-BIH dataset consists of annotated ECG signals of heartbeats from 5 significantly skewed classes, including Normal (N), Artial Premature (S), Premature ventricular contraction (V), Fusion of ventricular and normal (F), Fusion of paced and normal (Q), where the amount of instances from each class is [90589, 2779, 7236, 803, 8039]\cite{DBLP:conf/ichi/KachueeFS18}.
In this experiment, we first employ a Conv1d layer with kernel size 10 as an embedding operation to project the time serial scalars in sequences into hidden representations, and use the segments as the smallest units for modeling and explanation.
Since the reproduced version of ProSeNet often fails when training with class weights, we compare all baseline methods without class weights.

\begin{table}[tb]
	\centering
	\begin{small}
		\begin{tabular}{lcccc}
			\toprule
			Method   & Acc.              & Avg.P             & Avg.R             & AOPC              \\
			\midrule
			LSTM     & 0.985             & 0.923             & 0.851             & -                 \\
			CNN      & 0.984             & 0.931             & 0.900             & -                 \\
			\midrule
			SAN      & 0.940             & 0.734             & 0.722             & 0.176             \\
			ProSeNet & 0.978             & \textbf{0.938}    & 0.844             & 0.186             \\
			SESM     & \underline{0.982} & 0.931             & \underline{0.893} & \textbf{0.232}    \\
			SESM +w  & \textbf{0.987}    & \underline{0.932} & \textbf{0.921}    & \underline{0.230} \\
			\bottomrule
		\end{tabular}
	\end{small}
	\caption{Performance on MIT-BIH dataset. ``+w'' denote training with class weights. }
	\label{tab:ecg}
\end{table}

\begin{figure}[tb]

	\centering
	\begin{subfigure}[b]{0.8\linewidth}
		\centering
		\includegraphics[width=\linewidth]{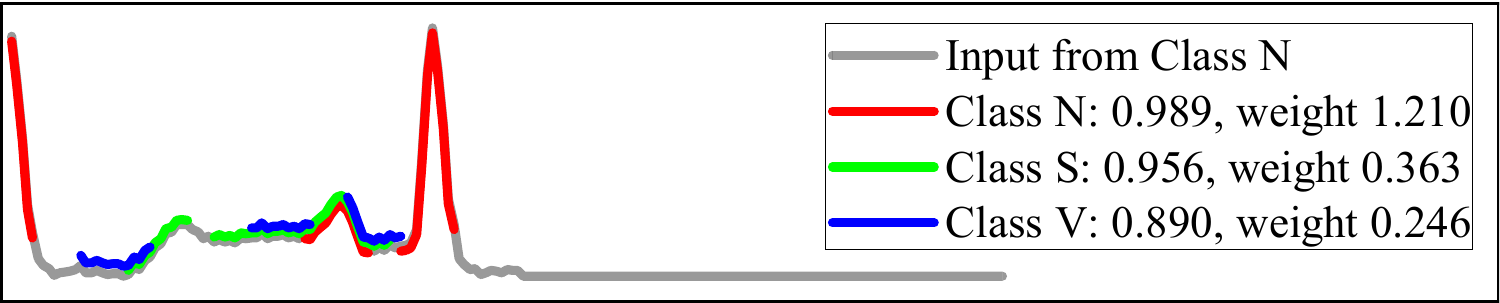}
	\end{subfigure}
	\hfill
	\begin{subfigure}[b]{0.8\linewidth}
		\centering
		\includegraphics[width=\linewidth]{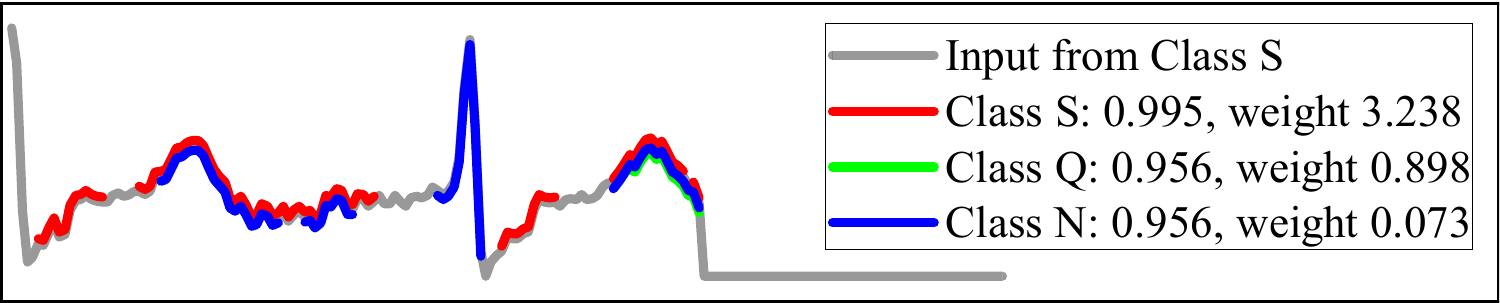}
	\end{subfigure}
	\hfill
	\begin{subfigure}[b]{0.8\linewidth}
		\centering
		\includegraphics[width=\linewidth]{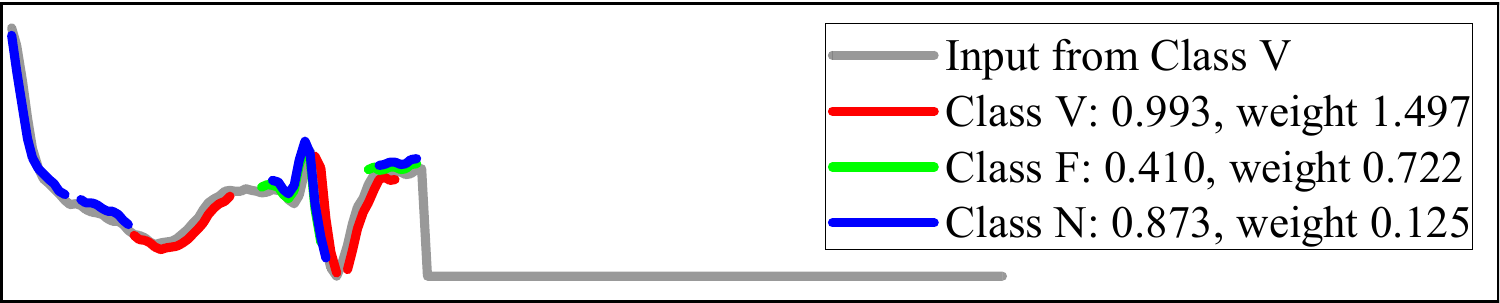}
	\end{subfigure}
	\hfill
	\begin{subfigure}[b]{0.8\linewidth}
		\centering
		\includegraphics[width=\linewidth]{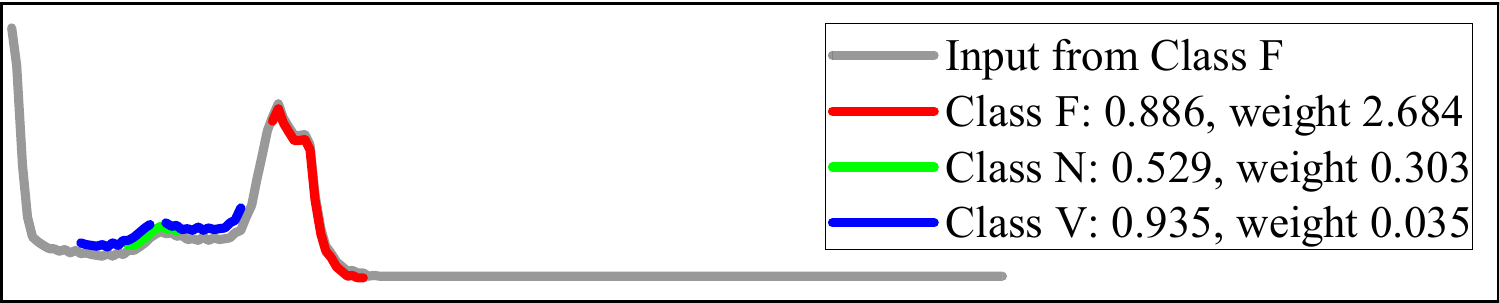}
	\end{subfigure}
	\hfill
	\begin{subfigure}[b]{0.8\linewidth}
		\centering
		\includegraphics[width=\linewidth]{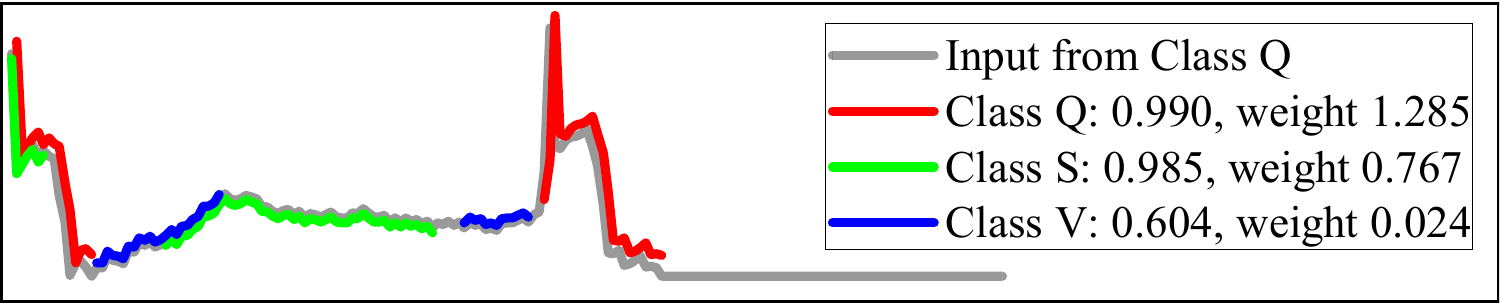}
	\end{subfigure}
	\caption{Case study on MIT-BIH dataset. Prototypical parts are marked with different colors on the input sequence. The legend shows the prediction probability based solely on the prototypical part, as well as the corresponding weight for aggregation.}
	\label{fig:ecg_case}
\end{figure}

\begin{figure}[tb]
	\centering
	\begin{subfigure}[b]{0.49\linewidth}
		\centering
		\includegraphics[width=\linewidth]{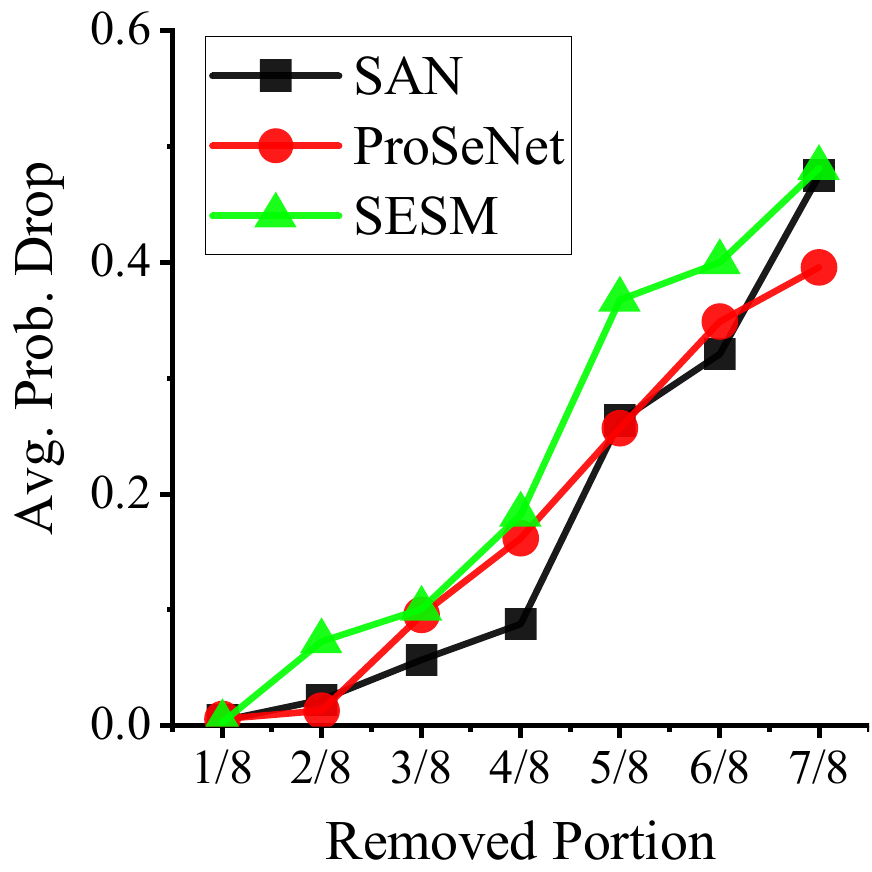}
		\caption{Counterfactual assessment. }
		\label{fig:aopc}
	\end{subfigure}
	\hfill
	\begin{subfigure}[b]{0.49\linewidth}
		\centering
		\includegraphics[width=\linewidth]{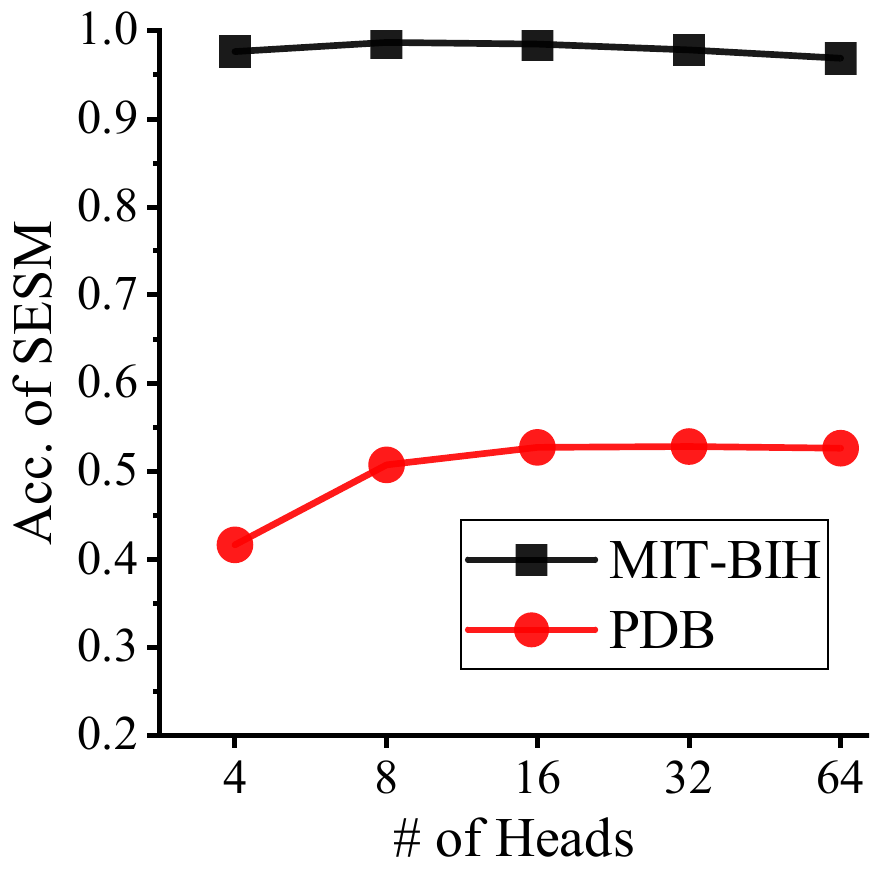}
		\caption{Impact of heads $H$. }
		\label{fig:heads}
	\end{subfigure}
	\hfill

	\caption{Detailed analysis on MIT-BIH dataset. }
	\label{fig:detail-ecg}
\end{figure}

\subsubsection{Task~2: Protein Family Classification}

The second task is protein family classification using PDB dataset\footnote{https://www.kaggle.com/shahir/protein-data-set} provided by Protein Data Bank.
The PDB dataset contains protein sequences in various lengths composed of 20 standard amino acids, and can be grouped into families.
We filtered out sequences whose length are less than 50 and clipped the sequences from the top 100 largest families with a maximal length of 512, following the settings of ProSeNet.

\begin{table}[tb]
	\centering
	\begin{small}
		\begin{tabular}{lcccc}
			\toprule
			Method   & Acc.              & Avg.P             & Avg.R             & AOPC              \\
			\midrule
			LSTM     & 0.526             & 0.381             & 0.333             & -                 \\
			CNN      & 0.538             & 0.399             & 0.322             & -                 \\
			\midrule
			SAN      & 0.517             & 0.380             & 0.311             & 0.130             \\
			ProSeNet & 0.489             & 0.334             & 0.277             & 0.092             \\
			SESM     & \textbf{0.526}    & \textbf{0.389}    & \textbf{0.333}    & \underline{0.135} \\
			SESM +w  & \underline{0.525} & \underline{0.381} & \underline{0.317} & \textbf{0.153}    \\
			\bottomrule
		\end{tabular}
	\end{small}
	\caption{Performance on PDB dataset.}
	\label{tab:protein}
\end{table}

\subsubsection{Task~3: Text Sentiment Classification}

The third task is text sentiment classification on the widely used dataset Yelp Reviews\footnote{https://www.yelp.com/dataset}, including binary classification (YelpP) and five-score classification (YelpF).
We tokenize the reviews with \texttt{Torchtext} and only use those with less than 25 words in the experiments.
The word embeddings are initialized with the 300D GloVe 6B pre-trained vectors \cite{DBLP:conf/emnlp/PenningtonSM14}.

\begin{figure*}[tb]
	\centering
	\includegraphics[width=.8\linewidth]{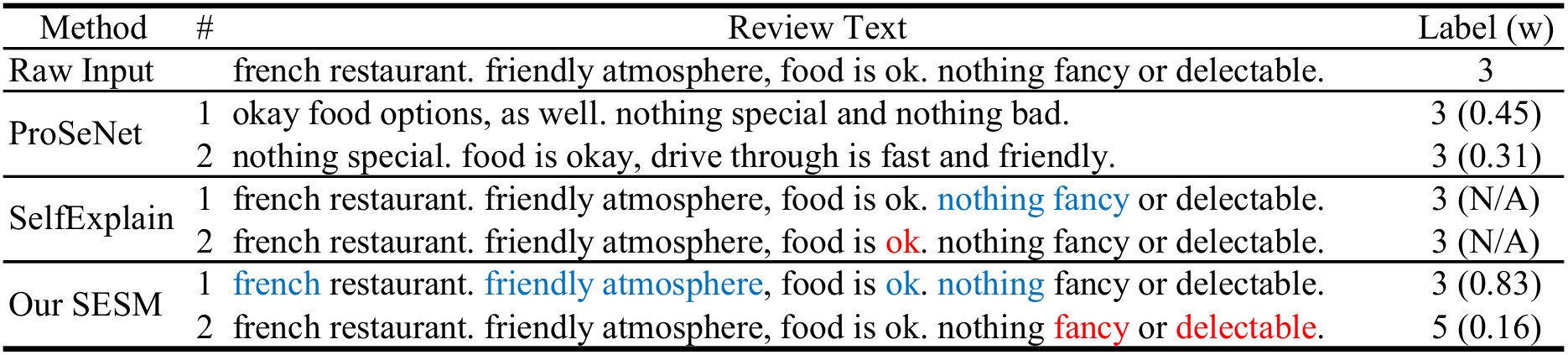}
	\caption{Case study on Yelp dataset. Colomn \# distinguishes different explanatory units. Color distinguishes words from different heads or sub-sequences.}
	\label{fig:yelp_case}
\end{figure*}

\begin{figure*}[tb]
	\centering
	\includegraphics[width=\linewidth]{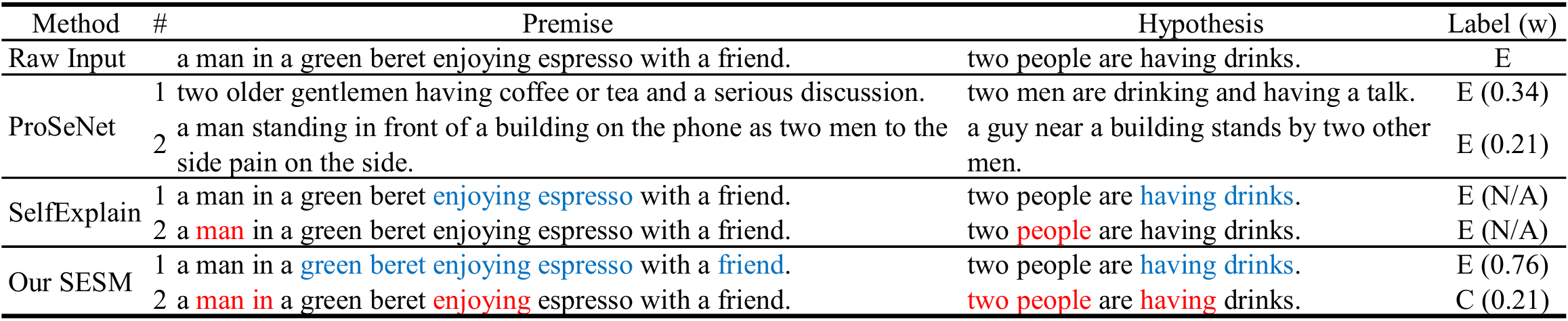}
	\caption{Case study on SNLI dataset.}
	\label{fig:snli_case}
\end{figure*}

\subsubsection{Task~4: Natural Language Inference}

We evaluate SESM on natural language inference task with the SNLI \cite{DBLP:conf/emnlp/BowmanAPM15} dataset, which aims to classify the semantic relation between a pair of sentences into three categories, including neutral (N), contradiction (C) and entailment (E).
We follow the experimental settings of existing work SAN \cite{DBLP:conf/iclr/LinFSYXZB17} and SSAN \cite{DBLP:conf/acl/GengWWQLT20}.
The sentences of premise and hypothesis are encoded by the model, denoted as $s_p$ and $s_h$ respectively.
Then, the classification is based on a multi-layer perceptron with two fully connected layers activated by $\operatorname{ReLU}$ on the concatenation of $[s_p, s_h, s_p-s_h, s_p\cdot s_h]$.

Since SelfExplain is originally designed as a classification method for single sentences, we tried to adapt SelfExplain to fit the pair-wise natural language inference task.
Specifically, the original SelfExplain encodes a complete sentence into a representation vector $s$, and find the hidden representations of concepts (words or phrases) that are most similar to $s$ for explanation.
By going through all possible pairs of concepts (generated by the constituency-based parse trees) from two sentences, we replace $s$ with $[s_p, s_h, s_p-s_h, s_p\cdot s_h]$, and the explanatory representations are then based on the most matchable pair-wise phrases.

\begin{table}[tb]
	\centering
	\begin{small}
		\begin{tabular}{clccc}
			\toprule
			Metric                & Method      & YelpP             & YelpF             & SNLI              \\
			\midrule
			\multirow{7}{*}{Acc.} & LSTM        & 0.953             & 0.704             & 0.786             \\
			                      & CNN         & 0.957             & 0.704             & 0.810             \\
			\cmidrule{2-5}
			                      & SAN         & \underline{0.952} & 0.692             & \underline{0.796} \\
			                      & ProSeNet    & \textbf{0.957}    & \underline{0.694} & 0.741             \\
			                      & SelfExplain & 0.951             & 0.689             & 0.728             \\
			                      & SESM        & 0.949             & \textbf{0.695}    & \textbf{0.803}    \\
			\midrule
			\multirow{4}{*}{AOPC} & SAN         & 0.363             & 0.257             & 0.267             \\
			                      & ProSeNet    & 0.375             & 0.222             & 0.208             \\
			                      & SelfExplain & \underline{0.379} & \underline{0.268} & \underline{0.287} \\
			                      & SESM        & \textbf{0.480}    & \textbf{0.336}    & \textbf{0.412}    \\
			\bottomrule
		\end{tabular}
	\end{small}
	\caption{Experimental results on NLP datasets.}
	\label{tab:result1}
\end{table}

\subsection{Experimental Results and Case Studies}

Tables~\ref{tab:ecg},~\ref{tab:protein}~and~\ref{tab:result1} show the experiment results on the accuracy and the quantified interpretability of models for comparison.
The \textbf{bold} values indicate the best results of the interpretable models for each entry, and the \underline{underlined} values indicate the second best results.
SESM shows comparable accuracy on various tasks, while all models with interpretability are facing performance loss to various degrees.
Nevertheless, SESM shows better interpretability on the counterfactual assessment AOPC, since it is designed to model disentangled concepts from different facets.

We illustrate the explanations on MIT-BIH dataset provided by SESM in Figure~\ref{fig:ecg_case}.
For each class, we randomly sample an instance of ECG signal, and highlight three most important prototypical parts representing different concepts that would be classified into distinct classes (i.e., skipping the sub-sequences that would lead to the same result).
Based on the selection of prototypical parts, SESM can directly mark the representative sub-sequences on the original input.

\begin{figure*}[tb]
	\centering
	\begin{subfigure}[b]{0.42\linewidth}
		\centering
		\includegraphics[width=\linewidth]{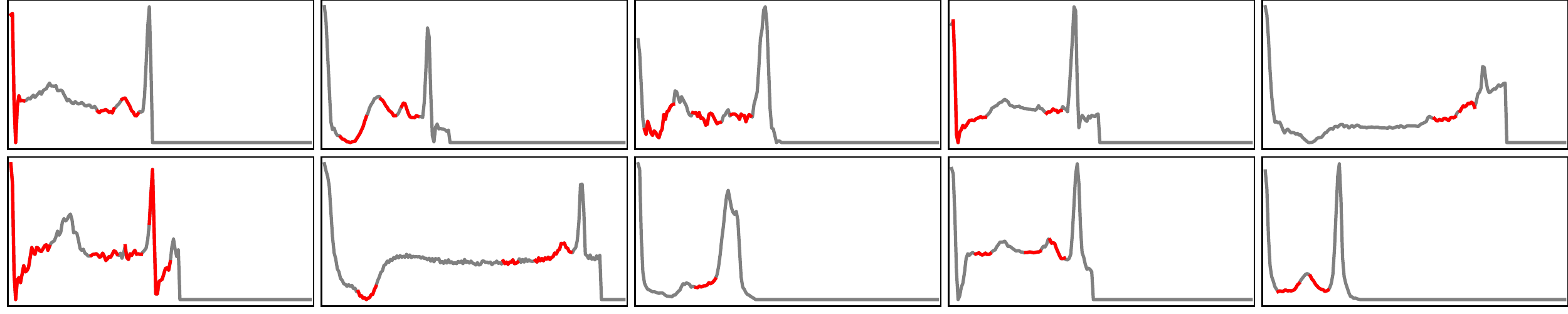}
		\caption{Head 1.}
		\label{fig:head1}
	\end{subfigure}
        ~ 
	\begin{subfigure}[b]{0.42\linewidth}
		\centering
		\includegraphics[width=\linewidth]{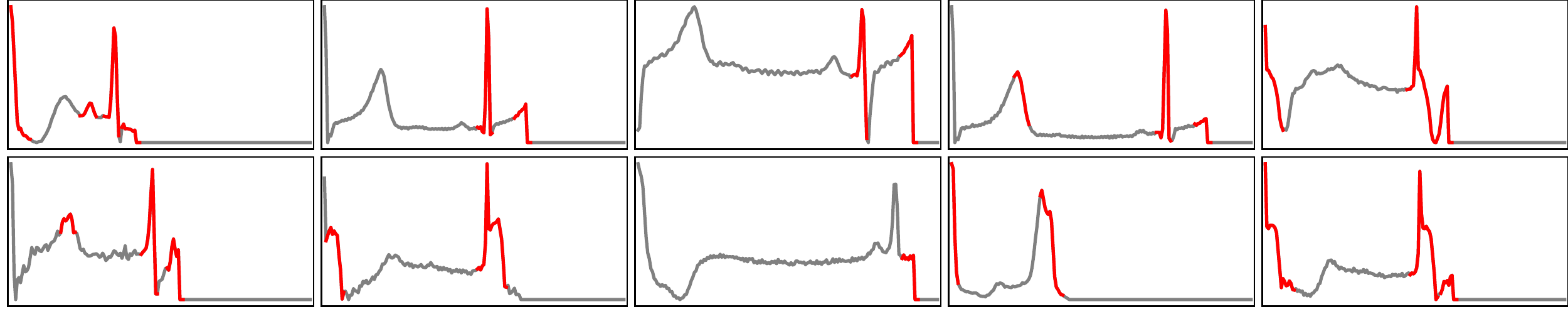}
		\caption{Head 2.}
		\label{fig:head2}
	\end{subfigure}
	\hfill
	\begin{subfigure}[b]{0.42\linewidth}
		\centering
		\includegraphics[width=\linewidth]{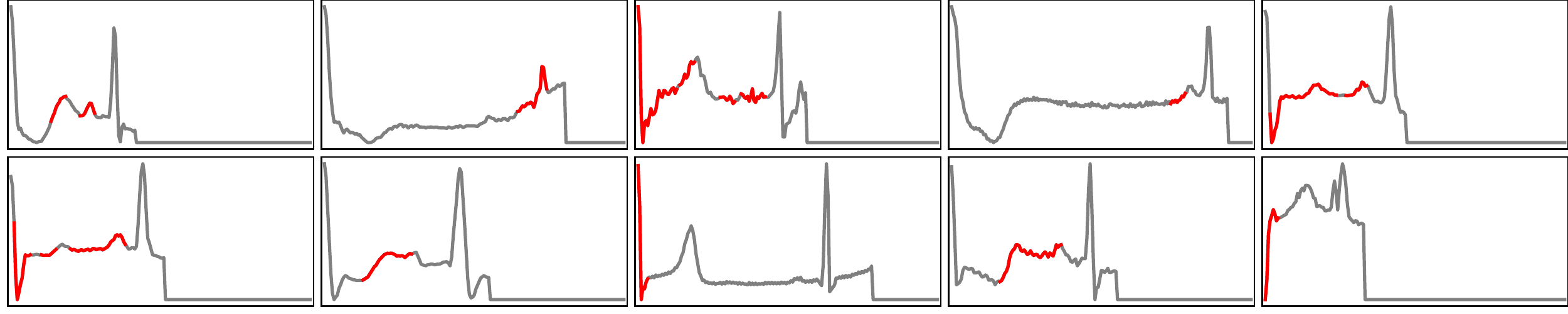}
		\caption{Head 3.}
		\label{fig:head3}
	\end{subfigure}
	~ 
	\begin{subfigure}[b]{0.42\linewidth}
		\centering
		\includegraphics[width=\linewidth]{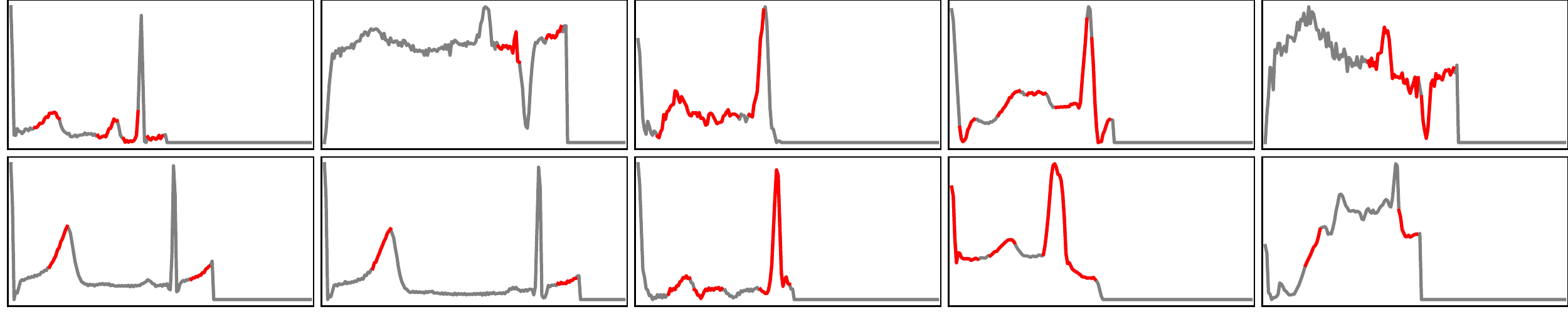}
		\caption{Head 4.}
		\label{fig:head4}
	\end{subfigure}
	\hfill
	\begin{subfigure}[b]{0.42\linewidth}
		\centering
		\includegraphics[width=\linewidth]{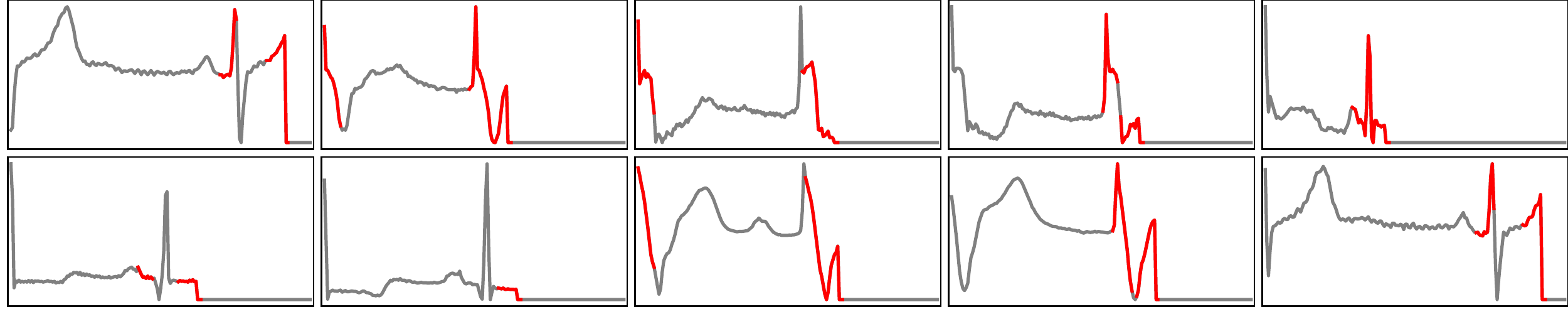}
		\caption{Head 5.}
		\label{fig:head5}
	\end{subfigure}
	~ 
	\begin{subfigure}[b]{0.42\linewidth}
		\centering
		\includegraphics[width=\linewidth]{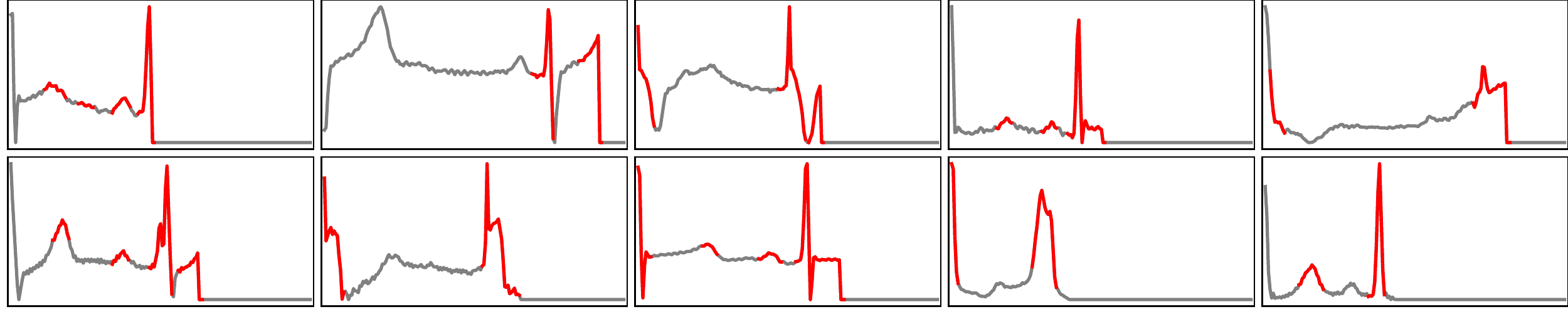}
		\caption{Head 6.}
		\label{fig:head6}
	\end{subfigure}
	\hfill
	\begin{subfigure}[b]{0.42\linewidth}
		\centering
		\includegraphics[width=\linewidth]{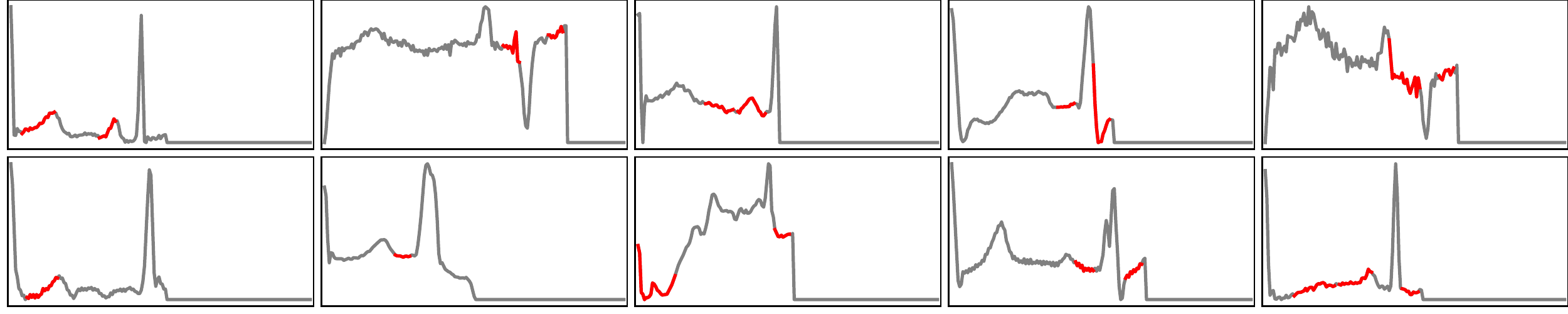}
		\caption{Head 7.}
		\label{fig:head7}
	\end{subfigure}
	~ 
	\begin{subfigure}[b]{0.42\linewidth}
		\centering
		\includegraphics[width=\linewidth]{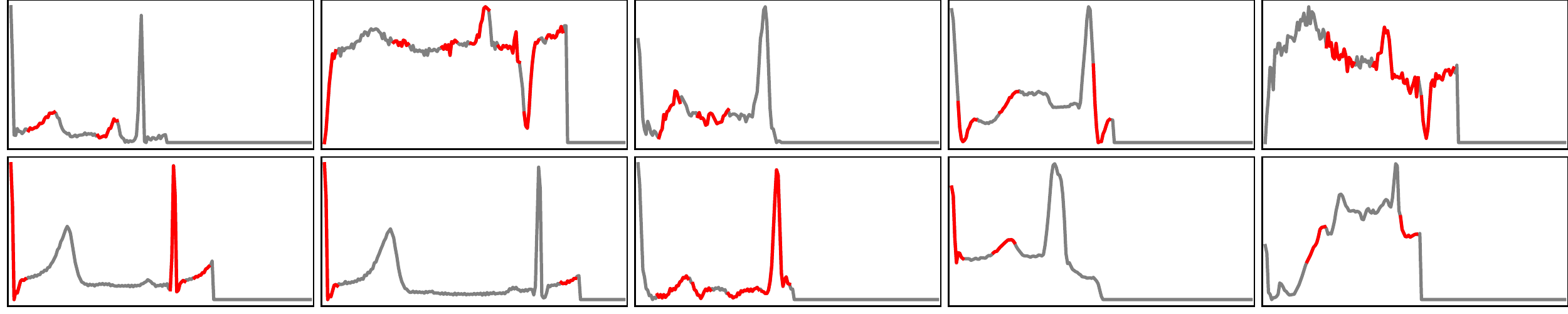}
		\caption{Head 8.}
		\label{fig:head8}
	\end{subfigure}
	\hfill
	\caption{Prototyping for explaining prototypical parts selected by different heads on MIT-BIH dataset. Note that the inputs are randomly sampled from different classes, but the selected prototypical parts activates the same concept within each head.}
	\label{fig:ecg_case_full}
\end{figure*}

\begin{figure}[tb]
	\centering
	\includegraphics[width=\linewidth]{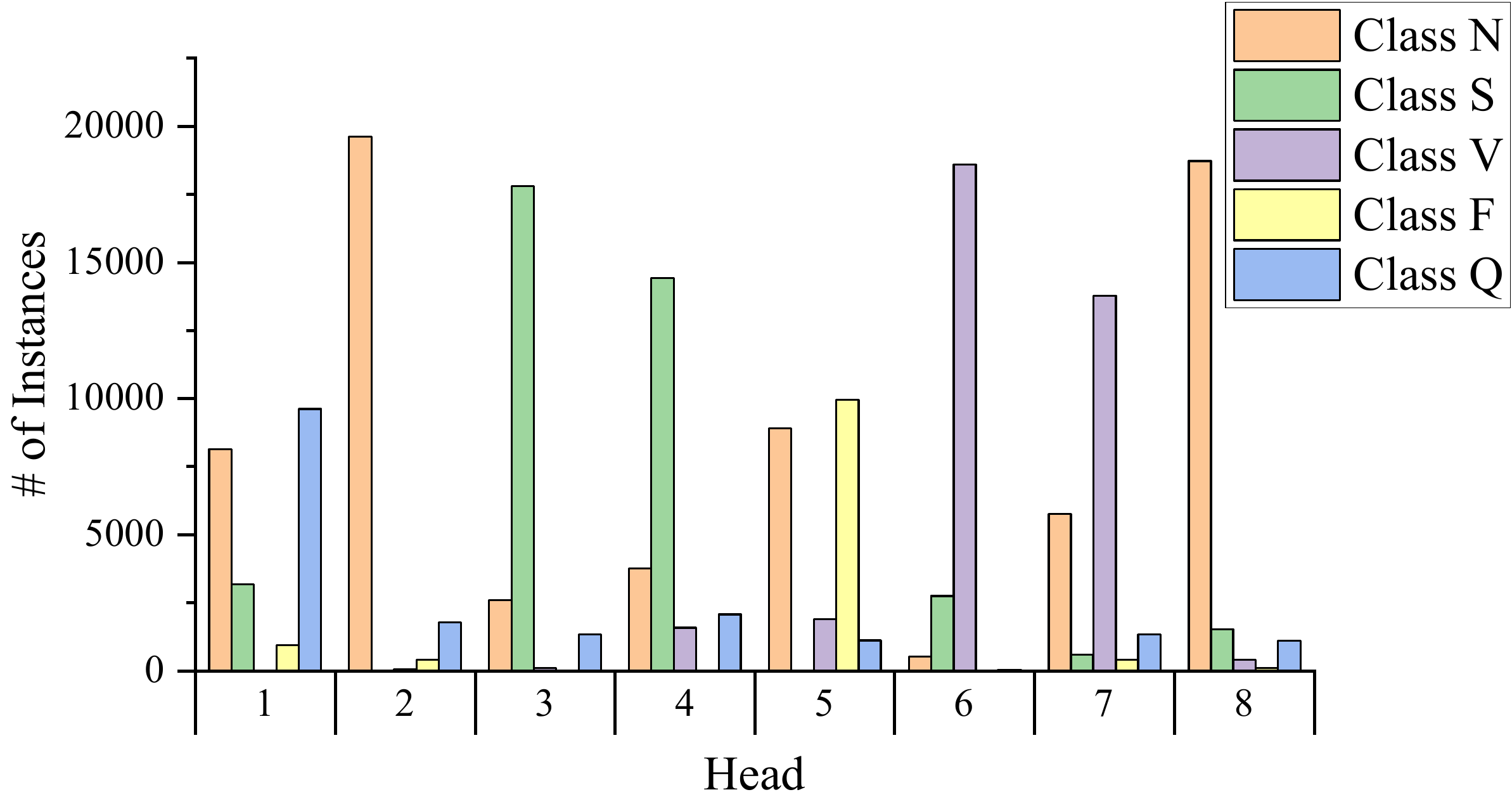}
	\caption{Statistics of the predictive results from different heads on MIT-BIH dataset. }
	\label{fig:ecg_stat}
\end{figure}

To better understand the explanations provided by different heads of SESM on MIT-BIH dataset, we illustrate the statistics of the predictive results in Figure~\ref{fig:ecg_stat} and instantiate 10 prototypical parts for each head in Figure~\ref{fig:ecg_case_full}.
The prototypes are provided to three laypersons of ECG diagnosis for human evaluation of model interpretability.
The average accuracy of human subjects increases from 0.543 to 0.691, which outperforms the situation of providing prototypes generated by ProSeNet (30 prototypes, increase from 0.543 to 0.610).
Presented below is a succinct overview of the inference process adopted by the three non-experts, upon being presented with the prototypical parts and the corresponding input for each head.

The inference process starts with the most distinguishable heads. According to Figure~\ref{fig:ecg_stat}, Heads~2, 3, 5 and 6 are with less ambiguity for the first four kinds of heartbeats N, S, F, and V, respectively.
As shown in Figure~\ref{fig:ecg_case_full}, Head~2 focuses on the crests of ECG signals, as well as Head~8. The selected prototypical parts illustrate how the crests arranged in normal heartbeats. Specifically, some minor bumps between two major peaks are typical for a normal heartbeat.
Head~3 generally attends to the pattern of abnormally elevation of waves just after the major peaks, which is clearly different from the N beats.
Analogously, the prototypical parts selected by Head~4 also shown the elevation of waves but at a lower degree, so that the predictive results of Head~4 is less certain.
Head~5 also focuses on the crests of ECG signals, but the selected prototypical parts are not as neat as those in Head~2, which are classified into Class F.
Head~6 shows the pattern of a relatively slow decline of the peaks. Combined with Head~7, the evidences for forecasting Class V are clear to grasp.
Finally, Head~1 simply attend to the pattern of a plateau followed by a cliff at the second peak as the evidence for Class Q.

Figures~\ref{fig:yelp_case}~and~\ref{fig:snli_case} show the case studies of NLP tasks on models with prototype-based interpretability. Compared with ProSeNet, SelfExplain and SESM provide concept-level prototypical parts as explanations, which is more intuitively understandable for laypersons. Compared with SelfExplain, SESM explicitly shows the weights of the prototypical parts for model prediction, instead of the post-hoc similarity-based analysis on the hidden representations. Besides, the prototypical parts used in SESM are more comprehensive than SelfExplain, since SESM imposes constraints that the selected prototypical parts should be distinct with each other, which also helps to provide counterfactual explanations about how to flip the model prediction by given only a sub-sequence of the input.

We further analyze the performance of SESM with different number of concepts, i.e. the number of heads $H$. Figure~\ref{fig:heads} shows that SESM performs better with 8 heads for MIT-BIH dataset and with 16 heads for PDB dataset, whereas it does not significantly affect the performance to keep increasing the number of heads. According to our further analysis, some of the excessive heads would tend to be consistent with each other, thus the diversity loss would increase but has a minor impact on the performance.

\section{Conclusion and Future Work} 

In this work, we propose a self-explaining selective model for interpretable sequence modeling named SESM, which selects sub-sequences from the raw input representing disentangled concepts as prototypical parts to provide more intuitively understandable explanations.
It eliminates discrepancies between hidden layer representation vectors and original sequence of prototypes when understanding prototype-based interpretations.
Experiments on various domains demonstrate that SESM consistently outperforms baseline interpretability methods with stronger impact on counterfactual assessment, while retaining comparable model accuracy with baseline methods with and without interpretability.
Future work includes designing an aggregator that could leverage the higher-order interaction of features from different concepts, while retaining interpretability.

\section*{Acknowledgments}

This work is supported by the National Key Research and Development Program of China.

\bibliography{aaai23}

\end{document}